\documentclass[10pt,twocolumn,letterpaper]{article}

\usepackage{cvpr}              

\usepackage{multirow}
\usepackage{threeparttable}
\usepackage{array}

\definecolor{cvprblue}{rgb}{0.21,0.49,0.74}
\usepackage[pagebackref,breaklinks,colorlinks,allcolors=cvprblue]{hyperref}

\def\paperID{*****} 
\def\confName{CVPR}
\def\confYear{2026}

\title{VISTA: Technical Report for the Ego4D Short-Term Object Interaction Anticipation at EgoVis 2026}

\author{
Qiaohui Chu$^{1\,2}$, Haoyu Zhang$^{1\,2}$, Yisen Feng$^{1}$, Meng Liu$^{3}$, Weili Guan$^{1}$, \\ Dongmei Jiang$^{2}$, Liqiang Nie$^{1}$\\
$^1$Harbin Institute of Technology (Shenzhen) \qquad  $^2$Pengcheng Laboratory    \\$^3$Shandong Jianzhu University\\
{\tt\small \{qiaohuichu8599, zhang.hy.2019, yisenfeng.hit, mengliu.sdu\}@gmail.com;} \\
{\tt\small \{honeyguan, nieliqiang\}@gmail.com; jiangdm@pcl.ac.cn}
}

\begin{document}
\maketitle
\begin{abstract}
We propose VISTA, a \textbf{V}-JEPA \textbf{I}ntegrated \textbf{S}tillFast \textbf{T}emporal \textbf{A}nticipator for the Ego4D Short-Term Object Interaction Anticipation (STA) Challenge at EgoVis 2026. 
Given an egocentric video timestamp, the task requires anticipating the next human-object interaction, including the future active object's bounding box, noun category, verb category, time-to-contact, and confidence score. 
VISTA follows a StillFast-style design that combines object-centric spatial detection with short-horizon temporal context. 
Specifically, a COCO-pretrained Faster R-CNN ResNet-50 FPN detector generates object proposals from the last observed high-resolution frame, while a frozen V-JEPA 2.1 temporal branch extracts clip-level egocentric context from the observed video. 
The temporal representation is injected into the detection pathway through feature modulation and ROI-level context fusion. 
The fused proposal features are then passed to multi-head STA predictors for box refinement, noun classification, verb classification, time-to-contact regression, and interaction confidence estimation. 
For the final submission, we further ensemble complementary predictions to improve robustness. 
Experimental results on the official challenge server show that VISTA achieves first place in the EgoVis 2026 Ego4D STA Challenge. Our code will be released at \url{https://github.com/CorrineQiu/VISTA}.

\end{abstract}    
\section{Introduction}
\label{sec:intro}

Egocentric video understanding has become an important research direction for embodied perception and human-assistive systems. 
Unlike third-person videos, egocentric videos are captured from the camera wearer's viewpoint and naturally record hands, gaze-driven scene layouts, object interactions, and user intentions. 
This perspective provides rich evidence for understanding human behavior and enables intelligent agents to offer proactive assistance. 
At the same time, egocentric videos are difficult to analyze. 
Frequent head and body movements introduce strong camera motion, active objects are often partially visible or occluded, and the relevant interaction cues may appear only briefly before contact. 
These challenges make egocentric anticipation substantially different from standard action recognition. Recent studies further show that egocentric video understanding often requires complementary abilities beyond frame-level recognition, including context-aware intention modeling, adaptive selection of informative visual evidence, cross-view knowledge transfer, and spatial reasoning. Context-aware multimodal reasoning has been explored through relational graph modeling for user intention understanding \cite{zhang2021multimodal}. In egocentric video, adaptive vision selection has been used to address small objects, noisy observations, and spatial-temporal reasoning in question answering \cite{pmlr-v235-zhang24aj}, while exocentric-to-egocentric knowledge transfer and structured spatial prompting further improve first-person video understanding and spatial reasoning \cite{zhang2026exo2ego,zhang2026spatial}. These observations motivate our design to combine object-centric spatial localization with temporal context for STA.

\begin{figure*}[t]
  \centering
  \includegraphics[width=\linewidth]{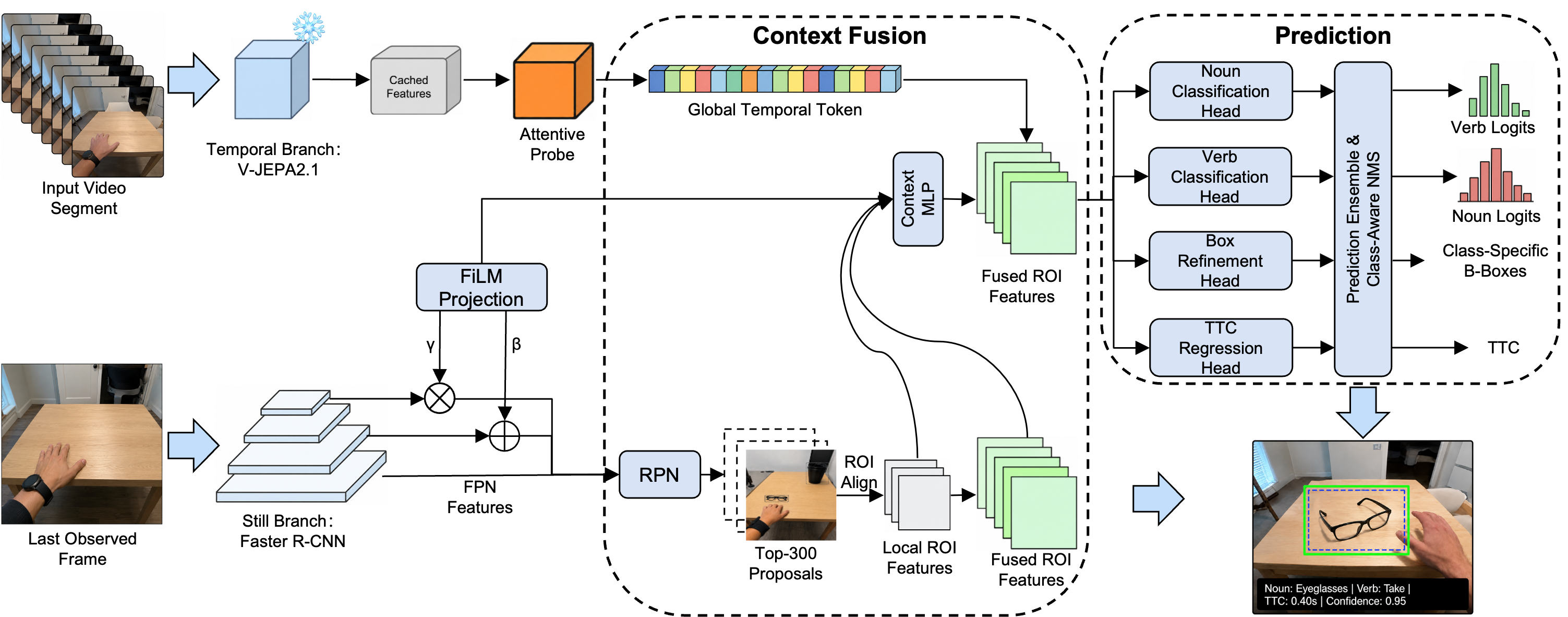}
  \caption{Overview of VISTA. VISTA combines a still-object detection branch with a frozen V-JEPA 2.1 temporal branch. Temporal context is fused into the FPN and ROI features before multi-head STA prediction and ensemble inference.}
  \label{fig:method}
\end{figure*}

The Ego4D Short-Term Object Interaction Anticipation
(STA) task, built on the Ego4D egocentric video dataset and benchmark suite \cite{grauman2022ego4d}, focuses on forecasting the next active object interaction after a given timestamp.
Given an untrimmed egocentric video and a timestamp $t$, a model can use frames observed up to $t$ and must output a set of future interaction hypotheses. 
Each hypothesis contains the bounding box of the future active object, the object's noun category, the future verb category, a time-to-contact (TTC) value indicating when the interaction will begin, and a confidence score. 
This task is important for assistive agents and human-machine collaboration, where anticipating imminent object contact can support early warning, robotic handover, and proactive assistance. Beyond short-term object interaction anticipation, Ego4D has also stimulated related egocentric tasks such as episodic-memory localization, long-term action anticipation, and long-form video question answering. Recent challenge solutions emphasize early fusion for temporal localization \cite{feng2025osgnet}, visual feature extraction with verb-noun co-occurrence and LLM-based future action prediction \cite{chu2025technical}, confidence-aware multi-source aggregation and fine-grained reasoning \cite{zhang2025hcqa}, and intention-guided cognitive reasoning based on hand-object interaction cues \cite{chu2026insight}. Compared with these tasks, STA requires each hypothesis to satisfy localization, noun, verb, and time-to-contact constraints simultaneously, making precise object-level temporal fusion especially important.

STA is challenging because it requires spatial localization, semantic prediction, and temporal forecasting at the same time. 
The model must localize a candidate object in the observed scene, infer which object will become active, predict the future interaction category before contact, and estimate the contact time within a short tolerance. 
Strong still-image detectors provide reliable object proposals, while temporal video representations help identify which visible object is likely to be involved in the next interaction. Recent baselines, such as Faster R-CNN with SlowFast and StillFast-style systems, show that combining spatial object evidence with temporal context is an effective direction \cite{feichtenhofer2019slowfast,ragusa2023stillfast,ren2015faster}. This direction is also consistent with recent egocentric VQA and episodic-memory studies, where task-relevant visual selection and early feature fusion are important for handling noisy first-person observations \cite{pmlr-v235-zhang24aj,feng2025osgnet}. However, the primary evaluation metric requires the box, noun, verb, and TTC predictions to be correct simultaneously. Thus, improving only one component is insufficient for strong STA performance.

To address this challenge, we propose VISTA, a \textbf{V}-JEPA \textbf{I}ntegrated \textbf{S}tillFast \textbf{T}emporal \textbf{A}nticipator. 
VISTA combines a COCO-pretrained Faster R-CNN ResNet-50 FPN still branch with a frozen V-JEPA 2.1 ViT-G temporal branch. 
The still branch provides object-centric localization and proposal features, while the temporal branch captures recent egocentric context from the observed clip. 
We inject the temporal representation into the still branch through feature modulation and ROI-level context fusion, enabling object proposals to be aware of recent hand-object dynamics. 
The fused representations are used to predict box refinements, nouns, verbs, TTC values, and interaction confidence. 
For the final submission, we train the model with the official training split and most validation annotations, and report the official test-set leaderboard result. 
VISTA achieves the best Overall Top-5 mAP on the EgoVis 2026 Ego4D STA Challenge leaderboard.

\section{Method}
\label{sec:method}

Figure~\ref{fig:method} illustrates the overall pipeline of VISTA. VISTA
combines a COCO-pretrained Faster R-CNN ResNet-50 FPN
still branch for object proposal generation with a frozen
V-JEPA 2.1 temporal branch for short-horizon egocentric
context modeling. The temporal context is injected into the
detection pathway through FPN-level feature modulation
and ROI-level context fusion. The fused proposal features
are used for multi-head STA prediction, followed by
ensemble inference for the final submission.

\begin{table*}[t]
  \centering
  \caption{Official test-set leaderboard for the Ego4D STA evaluation task. Overall Top-5 mAP is the primary ranking metric. Higher values indicate better performance.}
  \label{tab:leaderboard}
  \begin{threeparttable}
  \begin{tabular*}{\textwidth}{@{\extracolsep{\fill}}c l c c c c@{}}
    \toprule
    \# & Participant & Overall & Noun & Noun+Verb & Noun+TTC \\
    \midrule
    \textbf{1} & \textbf{corrine} & \textbf{5.40} & \textbf{27.26} & \textbf{16.15} & 8.95 \\
    2 & sun0710 & 5.13 & 23.83 & 14.52 & 8.07 \\
    3 & StillFast Baseline V2 & 5.12 & 25.06 & 13.29 & \textbf{9.14} \\
    4 & Faster R-CNN + SlowFast Baseline V2 & 3.61 & 26.15 & 9.45 & 8.69 \\
    \bottomrule
  \end{tabular*}
  \end{threeparttable}
\end{table*}

\subsection{Input Construction}

For each evaluated timestamp, we construct an egocentric
clip and a detection image. The video branch samples eight
observed frames at 2 FPS and resizes them to 384 pixels.
The still branch uses the last observed high-resolution frame
as the detection image. During training, images are resized
with short-side augmentation between 640 and 800 pixels,
while inference uses a short side of 800 pixels.

For the final challenge submission, we train the model with
the official Ego4D STA training split and most validation
annotations. Since the validation data are used to increase
the amount of supervised training data, we do not report
separate validation results. The final model is trained for
16 epochs with a learning rate of $1.0\times10^{-4}$, and
all quantitative results are reported on the official test split
evaluated by the challenge server.

\subsection{Still-Object Proposal Branch}

The still branch is based on a COCO-pretrained Faster
R-CNN ResNet-50 FPN detector~\cite{he2016deep,lin2014microsoft,lin2017feature,ren2015faster}.
It provides a ResNet-FPN feature hierarchy, RPN proposals,
ROIAlign features~\cite{he2017mask}, and a detector box
head. We replace the original COCO classification layer
with STA-specific prediction heads. This design preserves
the detector's generic object localization ability while
adapting the output space to Ego4D nouns, verbs, and
interaction timing.

At inference time, the detector keeps up to 300 candidate
proposals after RPN filtering. These proposals are not
collapsed into a single object at an early stage. Instead,
each proposal is encoded as a local object representation
and later ranked by the STA prediction heads. The ROI
features are pooled and processed by the detector box head,
producing proposal-level features for temporal context
fusion.

\subsection{Temporal Branch and Context Fusion}

The temporal branch uses a frozen V-JEPA 2.1 ViT-G
encoder at 384 resolution~\cite{mur2026vjepa}. To reduce
inference cost, we cache global V-JEPA features for
observed clips and keep the temporal branch frozen. A
lightweight attentive probe summarizes the cached feature
sequence into a global temporal token that represents recent
egocentric context.

We inject this temporal token into the still branch at two
levels. First, a FiLM-style projection~\cite{perez2018film}
generates feature-wise scale and bias terms to modulate FPN
features before proposal generation and ROI processing.
Second, the projected temporal token is concatenated with
each ROI feature and passed through a small context MLP.
The output residual is added to the local ROI representation.
This two-level fusion allows the detector to remain spatially
precise while making each object proposal aware of recent
hand-object interaction dynamics. This design is related to
adaptive visual evidence selection in egocentric VQA~\cite{zhang2024multi}
and early-fusion localization in Ego4D episodic-memory
tasks~\cite{feng2025osgnet}, but here the temporal context
is used to modulate object proposals for short-term
interaction anticipation.

\subsection{STA Prediction Heads}

The fused ROI features are passed to STA prediction heads.
For each candidate proposal, the model predicts noun logits
over Ego4D STA object classes, class-specific box
refinements, verb logits over future action classes, a
non-negative TTC value through a softplus regression head,
and an interaction quality score for ranking.

We train four prediction heads in parallel and average their
losses during optimization. At inference time, retained
proposals are expanded into top noun and verb hypotheses.
The final hypotheses are ranked by objectness, interaction
quality, noun probability, and verb probability, and are then
filtered with class-aware non-maximum suppression. The top
100 predictions are exported in the official STA submission
format.

\subsection{Prediction Ensemble}

The final submission uses an ensemble of complementary
predictions. The ensemble groups compatible hypotheses
according to noun category, verb category, box overlap,
and TTC proximity. Predictions within the same group are
reweighted and merged based on confidence and cross-head
agreement. This strategy reduces head- and checkpoint-level
variance while preserving the matching requirements of the
official STA metric. More broadly, our ensemble follows
the common idea of improving robustness by combining
complementary cues or predictions, which has also been
explored in collaborative learning and confidence-aware
multi-source reasoning settings~\cite{zhang2023attribute,zhang2025hcqa}.


\section{Experiments}
\label{sec:experiments}

\subsection{Evaluation Protocol}

Ego4D STA submissions are evaluated on the official test
split. The primary metric is Overall Top-5 mAP. A predicted
hypothesis can match a ground-truth interaction only when
the bounding-box IoU is greater than 0.5 and the required
semantic and temporal conditions are satisfied.

The challenge reports four Top-5 mAP variants. Noun mAP
requires the noun label to match. Noun+Verb mAP requires
both noun and verb labels to match. Noun+TTC mAP requires
the noun label to match and the TTC error to be less than
0.25 seconds. Overall mAP requires the box, noun, verb,
and TTC predictions to match simultaneously, and is used
as the primary ranking metric.


\subsection{Official Leaderboard Result}

Table~\ref{tab:leaderboard} reports the official test-set leaderboard for the STA evaluation task. Our submission, \textit{corrine}, achieves the best Overall Top-5 mAP of 5.40 and ranks first on the leaderboard. Compared with the StillFast Baseline V2, VISTA improves the primary Overall mAP from 5.12 to 5.40 and the Noun+Verb mAP from 13.29 to 16.15. Although the Noun+TTC score is slightly lower than the StillFast baseline, the improvement in the primary Overall metric indicates that VISTA better balances localization, semantic prediction, and temporal anticipation under the full matching criterion. These results show the effectiveness of combining still-image object detection, frozen V-JEPA 2.1 temporal context, context-aware ROI fusion, and prediction ensembling.

\subsection{Qualitative Examples}

\begin{figure}[t]
  \centering
  \includegraphics[width=\linewidth]{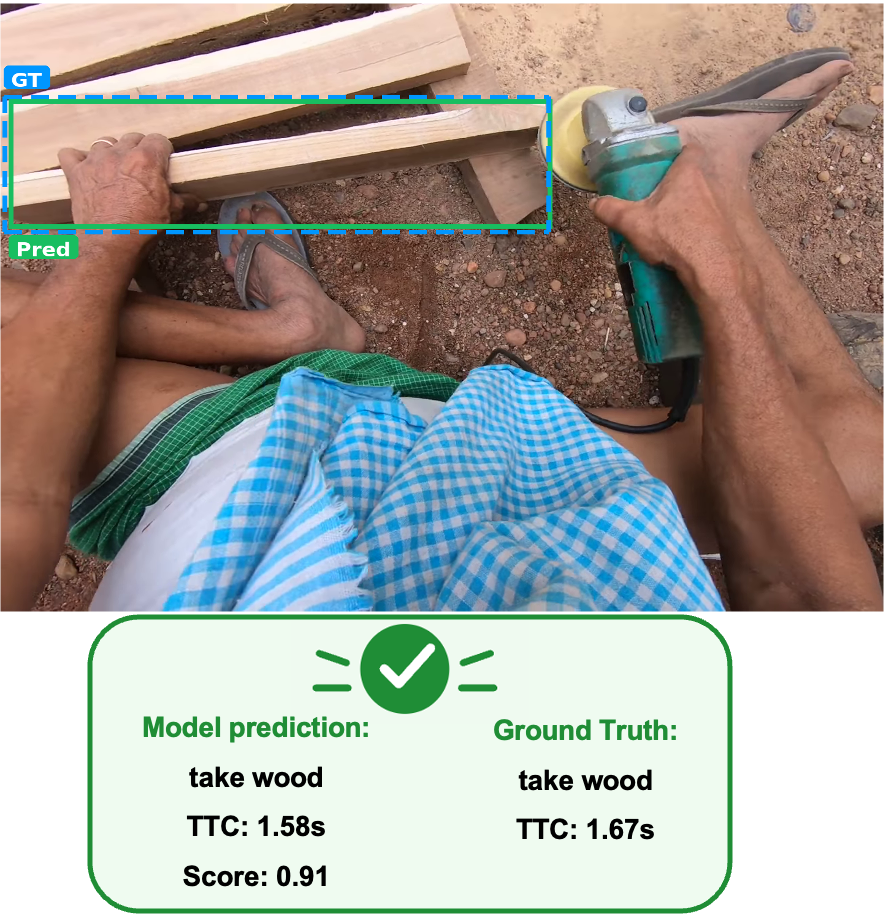}
  \caption{Successful qualitative example. VISTA correctly localizes the future active object and predicts the corresponding noun and verb.}
  \label{fig:success_case}
\end{figure}

\begin{figure}[t]
  \centering
  \includegraphics[width=\linewidth]{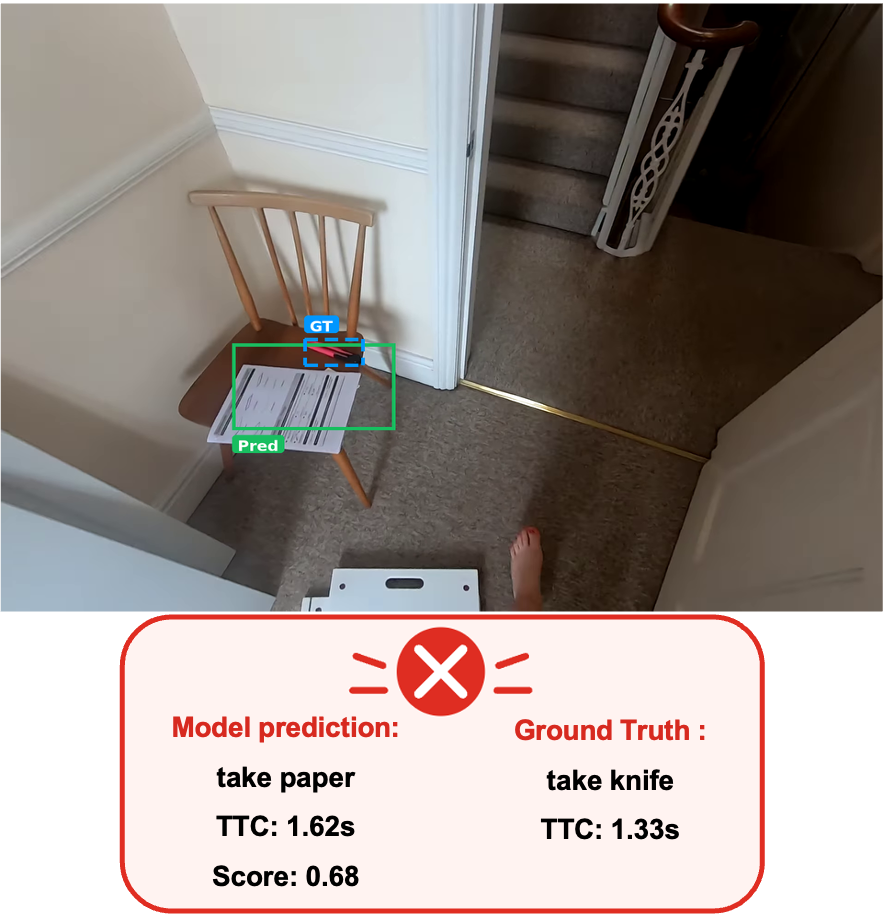}
  \caption{Failure qualitative example. VISTA is distracted by a visually salient paper region and misses the smaller ground-truth active object.}
  \label{fig:failure_case}
\end{figure}

Since the final model is trained with the official training split and most validation annotations, we do not report quantitative validation results. Instead, we provide two representative qualitative examples for diagnostic analysis. Figures~\ref{fig:success_case} and~\ref{fig:failure_case} show a successful case and a failure case, respectively, where solid green boxes denote model predictions and dashed blue boxes denote ground-truth active objects.

As shown in Figure~\ref{fig:success_case}, VISTA correctly anticipates the future interaction with the wooden plank, with accurate localization as well as correct noun and verb predictions. This example suggests that object-centric localization and temporal hand-object cues are complementary for STA. In contrast, Figure~\ref{fig:failure_case} shows that the model is distracted by a visually salient paper region and misses the smaller knife. This failure indicates that STA remains sensitive to small active objects and nearby distractors, suggesting that stronger hand-object region modeling and intention reasoning may further benefit STA \cite{pmlr-v235-zhang24aj,zhang2026spatial,chu2026insight}.
\section{Conclusion}
\label{sec:conclusion}

We presented VISTA, a V-JEPA Integrated StillFast Temporal
Anticipator for the EgoVis 2026 Ego4D Short-Term Object
Interaction Anticipation Challenge. VISTA combines a
COCO-pretrained Faster R-CNN ResNet-50 FPN still branch
with a frozen V-JEPA 2.1 temporal branch, and injects
temporal context into the detection pathway through
FPN-level feature modulation and ROI-level context fusion.
With multi-head STA prediction and ensemble inference,
VISTA achieves first place on the official challenge
leaderboard. These results show that combining
object-centric localization with frozen video representations
is effective for short-term object interaction anticipation.
{
    \small
    \bibliographystyle{ieeenat_fullname}
    \bibliography{main}
}


\end{document}